\documentclass[runningheads]{llncs}
\usepackage[T1]{fontenc}
\usepackage{graphicx}
\usepackage{pifont}
\usepackage{array}
\usepackage{multirow}
\usepackage{booktabs}

\newcolumntype{L}{>{\centering}m{1.5cm}}
\newcolumntype{R}{>{\centering}m{2.5cm}}
\newcolumntype{T}{>{\raggedright}m{2.5cm}}

\begin{document}
%

% Balancing Performance and Efficiency in Vision-Language Navigation Systems
% Efficient Vision-Language Fusion for Object Goal Navigation in Resource-Constrained Robotics
\title{Balancing Performance and Efficiency in Zero-shot Robotic Navigation}

%
% \titlerunning{Efficient Zero-shot Robotic Navigation}
\author{Dmytro Kuzmenko\inst{1}\orcidID{0009-0009-9296-1450} \and
Nadiya Shvai\inst{2}\orcidID{0000-0001-8194-6196}} %\and
\authorrunning{D. Kuzmenko and N. Shvai}
\institute{Department of Multimedia Systems, National University of Kyiv-Mohyla Academy, Kyiv, Ukraine
\\\email{kuzmenko@ukma.edu.ua}
% \url{http://www.springer.com/gp/computer-science/lncs} 
\and
{Department of Mathematics, National University of Kyiv-Mohyla Academy, \\ Kyiv, Ukraine\\
\email{n.shvay@ukma.edu.ua}}
}
\maketitle              % typeset the header of the contribution
\begin{abstract}
We present an optimization study of the Vision-Language Frontier Maps (VLFM) applied to the Object Goal Navigation task in robotics. Our work evaluates the efficiency and performance of various vision-language models, object detectors, segmentation models, and multi-modal comprehension and Visual Question Answering modules. Using the \textit{val-mini} and \textit{val} splits of Habitat-Matterport 3D dataset, we conduct experiments on a desktop with limited VRAM. We propose a solution that achieves a higher success rate (+1.55\%) improving over the VLFM BLIP-2 baseline without substantial success-weighted path length loss while requiring \textbf{2.3 times less} video memory. Our findings provide insights into balancing model performance and computational efficiency, suggesting effective deployment strategies for resource-limited environments.

% At least 3 at most 5 keywords
\keywords{Robotics \and Zero-Shot \and Object Goal Navigation  \and Computer Vision  \and Optimization.}
\end{abstract}

% \section{Related Work}
% \documentclass[runningheads]{llncs}
% \usepackage[T1]{fontenc}
% \usepackage{graphicx}

% \begin{document}

\section{Introduction}

In recent years, the field of robotic navigation has made significant strides, yet navigating to specific objects within complex and novel environments remains a challenging task. Object Goal Navigation (ObjectNav) tasks demand that a robot effectively locate and navigate to a designated object based on high-level semantic understanding rather than simple geometric cues. This capability is essential for applications such as domestic robots and autonomous delivery systems where robots must operate in dynamic and unfamiliar settings.

Previous research has leveraged a variety of methods to tackle the ObjectNav challenge. These include reinforcement learning (RL) approaches and learning from demonstrations \cite{ramrakhya2023pirlnav}, the use of frontier semantic policy \cite{yu2023frontier}, and versatile combinations of visual-textual methods with RL agents \cite{majumdar2022zson}. Zero-shot learning methods \cite{ChenLKGY23,zhou2023esc,10373065} have demonstrated the ability to generalize navigation tasks without extensive task-specific training. Models like CLIP on Wheels (CoW) \cite{gadre2023cows} and SemUtil \cite{ChenLKGY23} employ vision-language models (VLMs) and large language models (LLMs) to enhance navigation by providing contextual understanding and semantic inference capabilities.

Despite these advancements, existing approaches often face limitations with computational efficiency and the ability to handle a wide range of object categories. Methods relying on LLMs typically require significant computational resources and may not be feasible for deployment on resource-constrained platforms. Furthermore, the transformation of visual cues into text for semantic evaluation introduces additional layers of complexity and potential information loss.

In this work, we present an optimization study of the Vision-Language Frontier Maps (VLFM) \cite{yokoyama2024vlfm} applied to the Object Goal Navigation task. Building upon the VLFM framework, which integrates object detection, segmentation, and vision-language models, we aim to enhance both the quality of the results and reduce the resource allocation of video memory (VRAM) required for a real-world inference within a local workstation. We use \textit{val-mini} and \textit{val} splits of the Habitat-Matterport 3D dataset \cite{ramakrishnan2021habitat} in our research. Our experiments, conducted on a desktop with an NVIDIA RTX 3060 GPU with 12GB VRAM, explore various model configurations to identify strategies that balance performance and computational efficiency. By employing models such as CLIP-ViT-B/32 \cite{radford2021learning}, lighter YOLOv7 versions \cite{wang2023yolov7}, and nanoLLaVA \cite{nanollava}, we demonstrate the improvements in success rate and reduced VRAM requirements, suggesting ways of effective deployment strategies for resource-limited environments.

% \end{document}
% \documentclass[runningheads]{llncs}
% \usepackage[T1]{fontenc}
% \usepackage{graphicx}

% \begin{document}

\section{Methods}

\subsection{3D Indoor Spaces Dataset}
We utilized the Habitat-Matterport 3D (HM3D) dataset \cite{ramakrishnan2021habitat} -- the largest dataset of 3D indoor spaces comprising of 1,000 high-resolution 3D scans of building-scale spaces generated from real-world environments. For the architecture and optimal module composition search, we used a small \textit{val-mini} split of the dataset that contains 2 scenes and 30 episodes from HM3D. We ran the final inference tests on the full validation split. Episodes refer to specific instances of the navigation task within a scene, where the robot must navigate to a designated object goal. Each episode is defined by a starting position and a target object within the scene, providing a structured framework for evaluating navigation performance.

\subsection{RL Policy}
In our experiments, we adhere to the original reinforcement learning algorithm used in the previous work, namely Point Goal Navigation (PointNav) \cite{anderson2018evaluation}, though we plan to train our own RL agent in future work. After spin-initialization - rotating the body and camera to scope the landscape of the environment - the robot navigates towards either a frontier waypoint or a target object waypoint, based on the detection status of the target object. This approach leverages the egocentric depth image and the robot’s relative distance and heading towards the goal point, without relying on RGB images. The policy was previously trained using Variable Experience Rollout (VER) \cite{2022-Wijmans-SOLENER}.

\subsection{VLFM with VQA}
VLFM \cite{yokoyama2024vlfm}, a new state-of-the-art approach that we based our experiments on makes use of the three key components: a vision-language model that comprehends scenes semantically by calculating cosine similarity between textual scene description and visual embeddings; a detection model that outlines objects of interest in the scene (e.g. YOLOv7 \cite{wang2023yolov7} or GroundingDINO \cite{liu2023grounding}); and a segmentation model, MobileSAM, that identifies the contours of objects of interest. 

We decided to expand the complexity of the solution by adding another model - a multimodal Visual Question Answering (VQA) \cite{antol2015vqa} model. It proved to improve the performance and the average reward on "val-mini", though it was not as efficient on full validation split. VQA is used to confirm if the object's contours are correct visually.

\subsection{nanoLLaVA}
nanoLLaVA \cite{nanollava} is a not-yet-published 1.1-parameter VLM designed to run efficiently on edge devices. As a base LLM, it uses Quyen-SE-v0.1 of the Qwen-1.5 family \cite{bai2023qwen} with efficient Flash Attention module \cite{dao2022flashattention}. It uses LLaVA-family VLM   \cite{liu2023visual} backbone architecture with a SigLip vision tower \cite{zhai2023sigmoid}  for enhanced spatial comprehension. Unlike BLIP-2 \cite{li2023blip2}, which was used in the previous work, nanoLLaVA is restricted to generating visual, but not textual embeddings. LLaVA-family models do not have the native implementation of encoding text into embeddings yet. We have thus not experimented with nanoLLaVA as a fully-capable VLM and restrained ourselves to using it as a VQA module. We leave cosine similarity calculation with text and image embeddings produced by nanoLLaVa for future work. 

\begin{figure}[t]
\includegraphics[width=\textwidth]{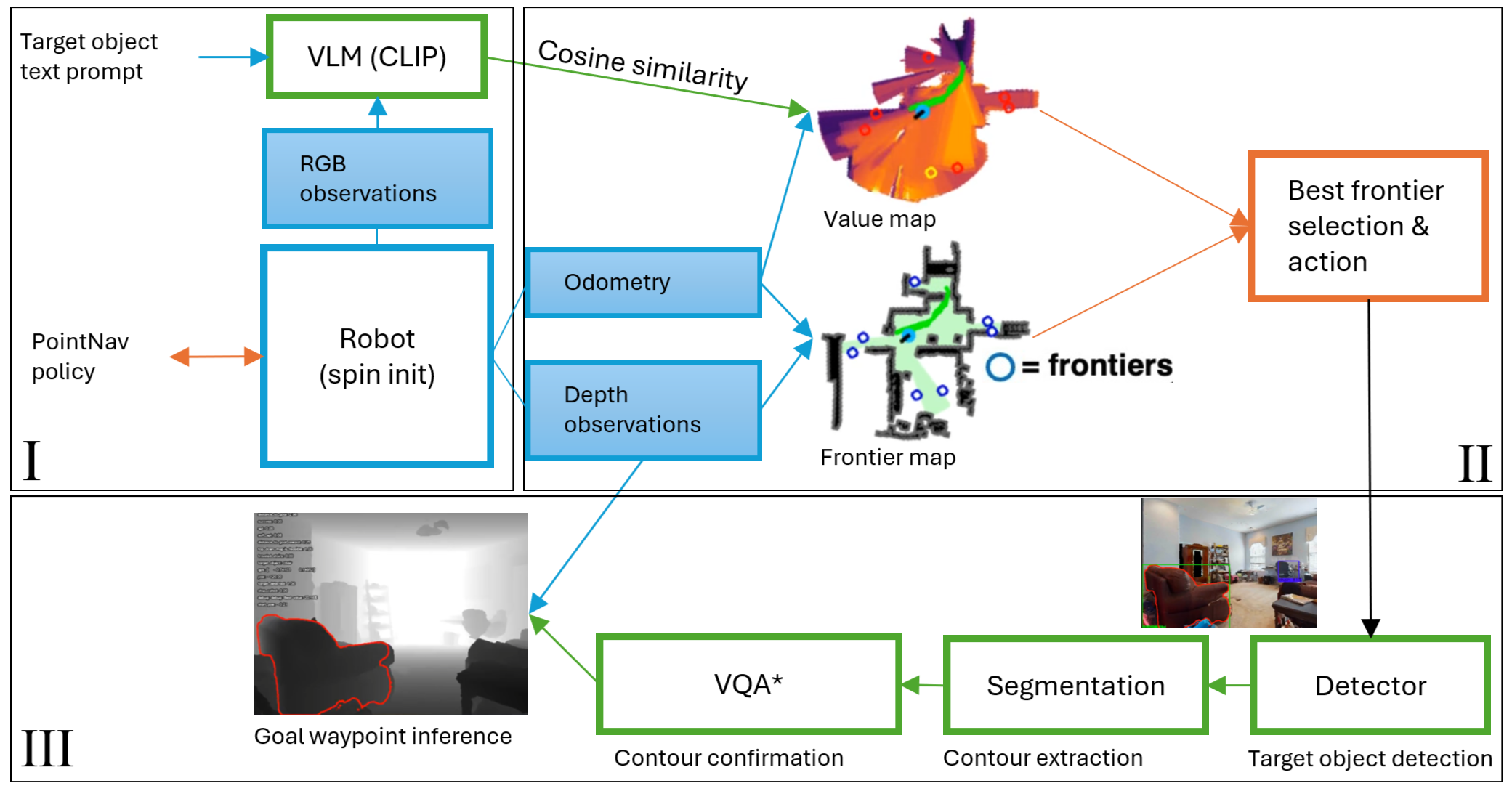}
\caption{A 3-stage diagram of our approach and its components. Before stage \textit{I}, all the models (green) are loaded and initialized. The decision-making components are marked in orange. Stage \textit{I} initializes the robot with RGB and depth cameras and the odometry (blue) and prepares the pre-trained RL policy; stage \textit{II} pre-computes cosine similarity and forms value and frontier maps based on color/depth observations and decides on an action; finally, stage \textit{III} processes the new scene with the object detector, segmentation model, and an optional VQA model. The resulting output is the inferred goal waypoint for the current step. \label{Figure 1}}
\end{figure}

The detailed diagram of our proposed approach with nanoLLaVA as VQA module is summarized in Figure \ref{Figure 1}.

\subsection{Model Parameter Study}
In our work, we outlined the parameter counts of all the models as these values directly influence the quality of the inference pipeline. We also denote the importance of allocated video memory (VRAM) used during the inference for resource-efficiency comparison (Table 1). The total VRAM allocation of a baseline VLFM approach is 6.494 MB; the total VRAM allocation of our final setup is 2.774 MB, approximately 2.3 times less.

The original VLM backbone, BLIP-2, utilizes Q-former \cite{li2023blip2} and CLIP-ViT-g \cite{radford2021learning} comprising 1.2B parameters. Our method, on the other hand, uses a simple CLIP-ViT-B32 with 151.3M parameters. As for the object detector, we replace YOLOv7-E6E (151.7M) with a lighter YOLOv7-W6 (70.4M) that has twice as few parameters \cite{wang2023yolov7}. MobileSAM \cite{zhang2023faster}, an efficient lightweight segmentation module, remains the consistent component from the original VLFM approach and has 9.8M parameters. 

Lastly, we add a new experimental component - a VQA module based on nanoLLaVA, which comprises 1.1B parameters and can be integrated into the pipeline owing to the replaced BLIP-2 and the freed-up VRAM, respectively.

\begin{table}[h]
    \centering
    \caption{Parameter count and VRAM requirements for the key models. Underlined models are the final choices for our lightweight approach.}
    \begin{tabular}{l|c|c}
        \hline
        \textbf{Model} & \textbf{Parameters, \#} & \textbf{VRAM required, MiB} \\
        \hline
        BLIP-2 & 1.4B & 2976 \\
        \underline{CLIP-ViT-B32} & 151.3M & \textbf{806} \\
        \hline
        YOLOv7-E6E & 151.7M & 3032 \\
        \underline{YOLOv7-W6} & 70.4M & \textbf{1482} \\
        \hline
        \underline{MobileSAM} & 9.8M & 486 \\
        \hline
        nanoLLaVA & 1.1B & 6002 \\
        \hline
        \hline
    \end{tabular}
    \label{tab:models}
\end{table}

% \end{document}
% \documentclass[runningheads]{llncs}
% \usepackage[T1]{fontenc}
% \usepackage{graphicx}
% \usepackage{pifont}
% \usepackage{array}

% \newcolumntype{L}{>{\centering}m{1.75cm}}
% \newcolumntype{R}{>{\centering}m{3cm}}

% \begin{document}

\section{Experiments}

\begin{figure}
\includegraphics[width=\textwidth]{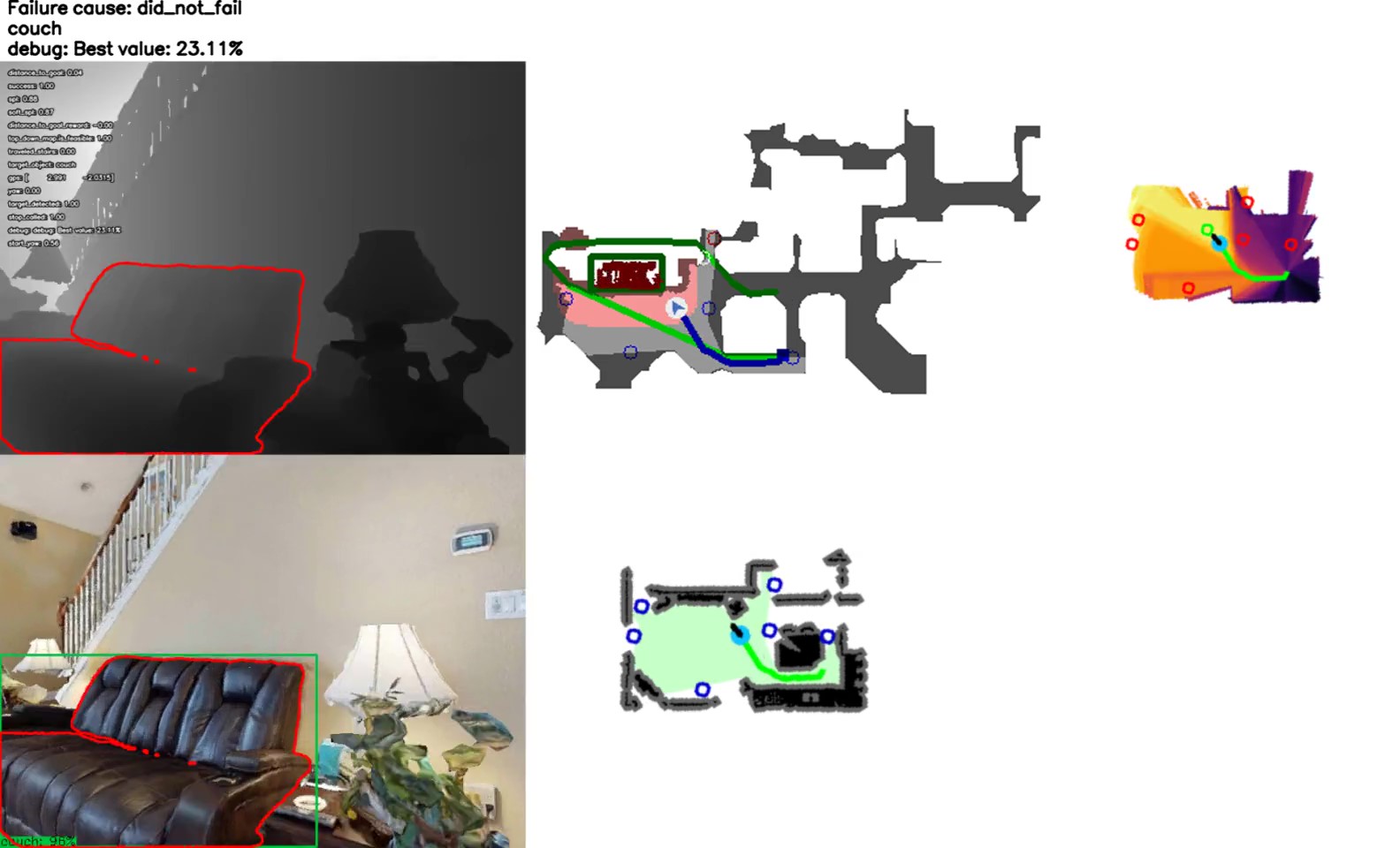}
\caption{A visualization of a frame at the successful end of the episode: top-left view illustrates the robot POV camera depth-view which is processed by segmentation/VQA models; bottom-left view showcases the results of object detection; the right-hand side demonstrates frontier maps and value maps.} \label{Figure 2}
\end{figure}

We conducted our experiments with the intent to both improve the system's performance and reduce the inference resource allocation requirement, making it better suited for real-world applications using a budget-oriented workstation. We focused on the key components of VLFM and looked to re-evaluate the pre-trained checkpoint with different VLMs, detectors, and by adding a VQA module. The robotic agent is equipped with odometry sensors as well as a depth camera and RGB camera for efficient environment navigation; with these three observation types, our model generates value maps and frontier maps that combine into a waypoint that guides the agent to the goal (Figure \ref{Figure 2}).

\subsection{Vision-Language Model}

The crucial part of the system architecture and design was the selection of the VLM that handles multimodal inputs and computes the cosine similarity. 

We initially wanted to use a compact and lightweight state-of-the-art model like nanoLLaVA as a main VLM. Aware of LLaVA's lack of text encoding implementation, and wanting to produce consistent embeddings of both modalities with the same model, we decided not to incorporate nanoLLaVA in the pipeline as a VLM.

Instead, we opted to explore the predecessor of BLIP-2 - the Contrastive Language-Image Pre-training (CLIP) model. The latter has different backbones, sizes, and weights to choose from. We used multiple backbones from open-clip \cite{ilharco_gabriel_2021_5143773} based on Vision Transformer (ViT) \cite{dosovitskiy2020image}, namely CLIP-ViT-B-32 and CLIP-ViT-B-16, and a larger ConvNext-XXL \cite{liu2022convnet} backbone in our experiments. 

\subsection{Object Detector}
The task of object detection involves determining if a target object is currently visible to the robot or not. The previous work uses the pre-trained largest YOLOv7-E6E \cite{wang2023yolov7} model to detect objects from COCO classes, and GroundingDINO \cite{liu2023grounding} -- to correctly identify other types of objects, namely the out-of-distribution samples. We re-use the previously suggested YOLOv7 but expand our experiments further to different YOLOv7 family models. 

We refrain from using GroundingDINO as we aim to make the system as lightweight and resource-efficient as possible to allow for the best real-time inference. We decided not to expand the number of classes that our detector module can detect at the cost of the fifth model in the solution. In our work, we try different YOLOv7 backbones, from most to the least lightweight, i.e. YOLOv7, YOLOv7-W6, YOLOv7-E6, YOLOv7-E6E.

\subsection{Segmentation Model}
A segmentation model extracts the contour of a successfully detected object using the RGB image and the bounding box. The depth image is then combined with the contour to identify the nearest point of the object relative to the robot's current position. This point is used as the goal waypoint for navigation. We keep this module of the pipeline unchanged and use lightweight MobileSAM \cite{zhang2023faster} for segmentation.

\subsection{VQA Module}
A VQA model is capable of processing the embeddings of two modalities, namely visual and textual. When provided with the bimodal input, the model responds coherently and accurately. Most modern VLMs \cite{chen2022pali,wang2022image,he2024malmm} are capable of high performance on VQA benchmarks. We decided that nanoLLaVA may be a good fit as a VQA model in our solution. Besides nanoLLaVA, we tried BLIP-2 proposed in the original work but were unable to fit in all the models in the memory restricted by 12 GB VRAM.

\subsection{Val-mini experiments}
As a main backbone, we used CLIP-ViT-B32 pre-trained on Laion2b \cite{schuhmann2022laion5b} with 34B tokes, YOLOv7-E6E as a starting detector, and MobileSAM as a segmentation model. We started with CLIP as a visual encoder, and DistillBERT \cite{sanh2020distilbert} and RoBERTa \cite{liu2019roberta} as textual encoders. These setups achieved little success as the embedding modality was different. It required additional linear transformation and implied extra information loss. Furthermore, the extractive power of older transformers may be insufficient when compared to the capabilities of modern LLMs and multimodal encoders. We then tried CLIP as a text-image encoder, which resulted in an improved success rate (+3.33\% compared to BLIP-2) and a faster inference time (-3.4 minutes) on \textit{val-mini}.

Next, we introduced nanoLLaVA to the pipeline enabling the VQA module. We ran tests with CLIP-ViT-B32, CLIP-ViT-B16, and ConvNext-XXL, with two former ones producing substantially weaker results and slower inference (Table 2).

\begin{table}[ht]
\centering
\caption{Our experiments on a val-mini split of HM3D dataset in zero-shot object goal navigation. * indicates the total time taken to infer all the modules for 30 episodes. ** denotes a baseline from VLFM inferred on val-mini.}
\label{tab:experiment_results}
\begin{tabular}{TlcLllc} % |T|l|c|L|l|l|c|
\hline
\textbf{VLM} & \textbf{Detector} & \textbf{VQA} & \textbf{Avg. reward} & \textbf{SPL$\uparrow$} & \textbf{SR$\uparrow$} & \textbf{Time, min*} \\ \hline
BLIP-2** & YOLOv7-E6E & - & 3.45 & \textbf{32.46} & 53.33 & 24.12\\ \hline
CLIP-ViT-B32 & YOLOv7-E6E & - & 3.5 & 28.2 & 56.67 & 20.71\\ 
CLIP-ViT-B32 + DistillBERT & YOLOv7-E6E & - & 3.87 & 27.74 & 53.33 & 27.53\\ 
CLIP-ViT-B32 + Roberta & YOLOv7-E6E & - & 3.8 & 27.85 & 50 & 28.21\\ 
ConvNext-XXL & YOLOv7-E6E & - & 3.00 & 25.75 & 43.33 & 28.59\\
CLIP-ViT-B16 & YOLOv7-E6 & \ding{51} & 3.55 & 25.91 & 56.67 & 29.66\\ 
CLIP-ViT-B32 & YOLOv7-E6E & \ding{51} & 4.54 & 28.37 & 63.33 & 26.47\\ 
CLIP-ViT-B32 & YOLOv7 & \ding{51} & 3.94 & 28.7 & 63.33 & 29.11\\
CLIP-ViT-B32 & YOLOv7-E6E & \ding{51} & 4.51 & 30.35 & \textbf{70} & 28.82\\ 
CLIP-ViT-B32 & YOLOv7-W6 & \ding{51} & \textbf{4.68} & \underline{29.42} & \textbf{70} & \underline{24.98} \\ 
\hline
\hline
\end{tabular}

\end{table}

\subsection{Inference Time Analysis}
At first, we assumed that using more lightweight modules would be sufficient to reduce the inference time by trading off some performance in return. However, as we proceeded with the experiments and measured the contribution of each module to the total inference time, we noticed that although lighter modules are faster, their less accurate predictions and lower-quality embeddings yield a much higher number of steps per episode, resulting in a longer runtime compared to baseline.

For example, a large BLIP-2 VLM takes 124 ms per inference step, in contrast to CLIP's 75 ms, but the average number of steps per episode is significantly reduced (hence fewer calls to the VLM and faster inference). The average number of steps is a derivative metric from success-weighted path length (SPL). Even though the setup is faster and has more lightweight components, it does not yet guarantee that the inferred semantic, value, and frontier maps as well as actions taken would be of high quality to outperform the more complex setup. Seeing this balance between the quality of the extracted features and the consistency in the agent's actions and their count, we aim to find an optimal middle-ground between the two. 

\begin{table}[h]
    \centering
    \caption{Performance metrics and inference speed results of different VLM and detector configurations on \textit{val-mini}.}
    \begin{tabular}{llcLLLcc} % |l|l|c|L|L|L|c|c|
        \hline
        \textbf{VLM} & \textbf{Detector} & \textbf{VQA} & \textbf{Avg. steps per episode} & \textbf{VLM infer., ms} & \textbf{Det. infer., ms} & \textbf{SPL$\uparrow$} & \textbf{SR$\uparrow$} \\
        \hline
        BLIP-2 & YOLOv7-E6E & - & \textbf{141} & 123.6 & 206.2 & \textbf{32.46} & 53.33 \\
        CLIP & YOLOv7-E6E & \ding{51} & 179 & 75 & 206.2 & 30.35 & \textbf{70} \\
        \textbf{CLIP} & \textbf{YOLOv7-W6} & \ding{51} & 186 & 75 & 167.6 & 29.42 & \textbf{70} \\
        CLIP & YOLOv7 & \ding{51} & 209 & 75 & 168.2 & 28.7 & 63.33 \\
        \hline
        \hline
    \end{tabular}
    \label{tab:performance_mini}
\end{table}

\begin{table}[h]
    \centering
    \caption{Original VLFM approach (w/ BLIP-2) and the proposed method (w/ CLIP) comparison: total time spent per each component on \textit{val-mini} split.}
    \begin{tabular}{l|c|c}
        \toprule
        \textbf{Model} & \multicolumn{2}{c}{\textbf{Total time spent, s}} \\
        \cmidrule(lr){2-3}
        & \textbf{w/ BLIP-2} & \textbf{w/ CLIP} \\
        \midrule
        VLM & 521.0 & 419.3 \\
        Detector & 850.4 & 937.7 \\
        Segmentation & 57.3 & 36.6 \\
        VQA & - & 105.2 \\
        \bottomrule
    \end{tabular}
    \label{tab:time_spent}
\end{table}

In Table \ref{tab:performance_mini}, we present the results of four evaluation runs on the \textit{val-mini} subset. It is apparent that with a weaker detector baseline, and thus the poorer detection quality, the agent is forced to take more steps on average throughout each episode, hindering SPL results, despite having faster inference time per step. We want to stress that it was not our goal to maximize both SR and SPL. Instead, we focused on SR and resource-efficiency of the pipeline with affordable hardware and accepted the fact that some SPL quality may be lost. We also analyzed the contribution of each component to the total inference time (Table 4).

In our experiments, we outlined the model complexity versus the steps per episode trade-off. The inaccuracies introduced by integrating more lightweight models increase the number of steps per episode. 

% TODO: references
\begin{table}[h]
\centering
\caption{Comparison of state-of-the-art approaches on the validation subset of HM3D dataset.}
\begin{tabular}{lRccc}
\hline
\textbf{Approach} & \textbf{Semantic Nav Training} & \textbf{Params, \#} & \textbf{SPL$\uparrow$} & \textbf{SR$\uparrow$} \\ \hline
PIRLNav & ObjectNav & $\sim$23M & 27.1 & \textbf{64.1} \\ 
ZSON & ImageNav & $\sim$85M & 12.6 & 25.5 \\ 
ESC & None & $\sim$65M & 22.3 & 39.2 \\ 
VLFM & None & 1.56B & \textbf{30.4} & 52.5 \\ \hline
VLFM with CLIP \& VQA (ours) & None & 1.33B & 27.24 & 53.35 \\ 
VLFM with CLIP (ours) & None & 231.5M & \underline{29.42} & \underline{54.05} \\ 
\hline
\hline
\end{tabular}
\end{table}

\section{Results}
Combining CLIP-ViT-B32 with nanoLLaVA produced different results for different detector backbones: the most lightweight YOLOv7 resulted in a weaker SPL (-3.76), and better success rate (+10\%), taking 20\% longer to infer due to the increased number of steps caused by inaccuracies and prolonged episode length; YOLOv7-W6, a medium-sized compact model yielded lower SPL (-3.04\%) and a substantial increase in success rate (+16.67\%) at the cost of 3.5\% slower inference speed. YOLOv7-E6, a slightly bigger model, did not yield any significant results. The default choice of the authors of VLFM, YOLOv7-E6E, proved to be a powerful alternative to YOLOv7-W6 despite its largest size -- highest SPL achieved (-2.04), a solid success rate of 70\%, with a considerably slower runtime, however.

We additionally evaluated our approach, with and without a VQA module, on 2000 episodes of the HM3D validation set for a fair comparison to other state-of-the-art models (Table 5). This ablation disproves the gains and benefits of a VQA model in more diverse and scaled-up distributions. The final models consist of CLIP-ViT-B32, YOLOv7-W6, and MobileSAM. The VLFM with CLIP \& VQA includes a nanoLLaVA that corrects the contour detected by the segmentation model. The former approach without VQA proved to be both faster (+2.18\% SPL compared to VQA variant) and more accurate (+0.7\% SR). When compared to VLFM, our model has a higher SR (+1.55\%), yet slightly lower SPL (-0.98\%). It is worth noting that our model's SPL is still higher than those of other previous SOTA approaches. 

The important gain of our work is the achievement of better results without a noticeable loss of inference speed, while also using 2.3 times less VRAM when compared to the original BLIP-2 setup.

% \end{document}
% \documentclass[runningheads]{llncs}
% \usepackage[T1]{fontenc}
% \usepackage{graphicx}

% \begin{document}

\section{Discussions}

In this work, we conducted an optimization study of the Vision-Language Frontier Maps applied to the Object Goal Navigation task in robotics. Our primary focus was on enhancing the efficiency and performance of the pipeline by optimizing its textual and visual components. Using the validation split of the Habitat-Matterport 3D dataset, we highlighted the effectiveness of models like CLIP-ViT-B32 with YOLOv7-W6 demonstrating significant improvements in success rate and negligible SPL loss compared to the baseline BLIP-2 model. We also disproved the high-scale efficiency of the VQA module based on nanoLLaVA, at least under current experimental scenarios.

Our results showed that the CLIP-ViT-B32 with YOLOv7-W6 achieved a 1.55\% higher SR, albeit with a slightly worse SPL (-0.98) compared to BLIP-2 VLFM. Compared to the BLIP-2 baseline, which has 1.4B parameters and requires 6494 MB VRAM, the CLIP-ViT-B32 model is significantly more compact with just 151.3M parameters and needs 2774 MB VRAM, which makes it 2.3 times more resource-efficient in terms of resource-allocation.

However, our research has several limitations. The optimized models have not been tested on real-world robots, limiting the practical applicability of our findings. Furthermore, our evaluations were confined to the HM3D dataset only. More comprehensive testing on diverse datasets like MP3D \cite{8374622} and Gibson \cite{xia2018gibson} is necessary to make our system better-generalized. Another limitation is that we did not modify or retrain the reinforcement learning policy (PointNav), which could further enhance navigation performance if optimized.

Future research should address these limitations by conducting real-world robot evaluations and expanding testing to larger validation sets. Additionally, refining system design and model selection processes can lead to further improvements. One promising direction is implementing a nanoLLaVA text embedding method to utilize the model as a vision-language model (VLM) with accurate cosine similarity computation. Using more diverse evaluation datasets and retraining the RL policy to better align with our optimized models is another step for future work, potentially unlocking higher efficiencies in object goal navigation tasks.

% \end{document}

% \begin{credits}
\subsubsection{\ackname} This research is dedicated to the people of Ukraine in response to the 2022 russian invasion and war. 
% \end{credits}

% \bibliographystyle{plain}
\bibliographystyle{splncs04}  % unsrt ieeetr


\begin{thebibliography}{10}
\providecommand{\url}[1]{\texttt{#1}}
\providecommand{\urlprefix}{URL }
\providecommand{\doi}[1]{https://doi.org/#1}

\bibitem{ramrakhya2023pirlnav}
Ramrakhya, R., Batra, D., Wijmans, E., Das, A.: {PIRLNav: Pretraining with
  Imitation and RL Finetuning for ObjectNav}. In: Proceedings of the IEEE/CVF
  Conference on Computer Vision and Pattern Recognition. pp. 17896--17906
  (2023)

\bibitem{yu2023frontier}
Yu, B., Kasaei, H., Cao, M.: {Frontier Semantic Exploration for Visual Target
  Navigation}. In: 2023 IEEE International Conference on Robotics and
  Automation (ICRA). pp. 4099--4105. IEEE (2023)

\bibitem{majumdar2022zson}
Majumdar, A., Aggarwal, G., Devnani, B., Hoffman, J., Batra, D.: {ZSON:
  Zero-Shot Object-Goal Navigation using Multimodal Goal Embeddings}. Advances
  in Neural Information Processing Systems  \textbf{35},  32340--32352 (2022)

\bibitem{ChenLKGY23}
Chen, J., Li, G., Kumar, S., Ghanem, B., Yu, F.: How to not train your dragon:
  Training-free embodied object goal navigation with semantic frontiers. In:
  Robotics: Science and Systems XIX, Daegu, Republic of Korea, July 10-14, 2023
  (2023). \doi{10.15607/RSS.2023.XIX.075}

\bibitem{zhou2023esc}
Zhou, K., Zheng, K., Pryor, C., et~al.: Esc: Exploration with soft commonsense
  constraints for zero-shot object navigation. In: Proceedings of the 40th
  International Conference on Machine Learning. ICML'23 (2023)

\bibitem{10373065}
Dorbala, V.S., Mullen, J.F., Manocha, D.: {Can an Embodied Agent Find Your
  “Cat-shaped Mug”? LLM-Based Zero-Shot Object Navigation}. IEEE Robotics
  and Automation Letters  \textbf{9}(5),  4083--4090 (2024).
  \doi{10.1109/LRA.2023.3346800}

\bibitem{gadre2023cows}
Gadre, S.Y., Wortsman, M., Ilharco, G., Schmidt, L., Song, S.: {Cows on
  pasture: Baselines and benchmarks for language-driven zero-shot object
  navigation}. In: Proceedings of the IEEE/CVF Conference on Computer Vision
  and Pattern Recognition. pp. 23171--23181 (2023)

\bibitem{yokoyama2024vlfm}
{Naoki Yokoyama and Sehoon Ha and Dhruv Batra and Jiuguang Wang and Bernadette
  Bucher}: Vlfm: Vision-language frontier maps for zero-shot semantic
  navigation. In: International Conference on Robotics and Automation (ICRA)
  (2024)

\bibitem{ramakrishnan2021habitat}
Ramakrishnan, S.K., Gokaslan, A., Wijmans, E., Maksymets, O., Clegg, A.,
  Turner, J., Undersander, E., Galuba, W., Westbury, A., Chang, A.X., et~al.:
  {Habitat-Matterport 3D Dataset ({HM}3D): 1000 Large-scale 3D Environments for
  Embodied AI}. arXiv preprint arXiv:2109.08238  (2021)

\bibitem{radford2021learning}
Radford, A., Kim, J.W., Hallacy, C., Ramesh, A., Goh, G., Agarwal, S., Sastry,
  G., Askell, A., Mishkin, P., Clark, J., et~al.: {Learning Transferable Visual
  Models From Natural Language Supervision}. In: International conference on
  machine learning. pp. 8748--8763. PMLR (2021)

\bibitem{wang2023yolov7}
Wang, C.Y., Bochkovskiy, A., Liao, H.Y.M.: {YOLOv7}: Trainable bag-of-freebies
  sets new state-of-the-art for real-time object detectors. In: Proceedings of
  the IEEE/CVF Conference on Computer Vision and Pattern Recognition (CVPR)
  (2023)

\bibitem{nanollava}
Nguyen, Q.: nanollava - sub 1b vision-language model (last accessed
  2024/05/26), \url{https://huggingface.co/qnguyen3/nanoLLaVA}

\bibitem{anderson2018evaluation}
Anderson, P., Chang, A., Chaplot, D.S., et~al.: On evaluation of embodied
  navigation agents. arXiv preprint arXiv:1807.06757  (2018)

\bibitem{2022-Wijmans-SOLENER}
Wijmans, E., Essa, I., Batra, D.: Ver: Scaling on-policy rl leads to the
  emergence of navigation in embodied rearrangement. In: Advances in Neural
  Information Processing Systems (NeurIPS) (2022).
  \doi{10.48550/arXiv.2210.05064}

\bibitem{liu2023grounding}
Liu, S., Zeng, Z., Ren, T., et~al.: {Grounding DINO: Marrying DINO with
  Grounded Pre-Training for Open-Set Object Detection}. arXiv preprint
  arXiv:2303.05499  (2023)

\bibitem{antol2015vqa}
Antol, S., Agrawal, A., Lu, J., Mitchell, M., Batra, D., Zitnick, C.L., Parikh,
  D.: {VQA: Visual question answering}. In: Proceedings of the IEEE
  international conference on computer vision. pp. 2425--2433 (2015)

\bibitem{bai2023qwen}
Bai, J., Bai, S., Chu, Y., et~al.: Qwen technical report. arXiv preprint
  arXiv:2309.16609  (2023)

\bibitem{dao2022flashattention}
Dao, T., Fu, D., Ermon, S., Rudra, A., R{\'e}, C.: Flashattention: Fast and
  memory-efficient exact attention with io-awareness. Advances in Neural
  Information Processing Systems  \textbf{35},  16344--16359 (2022)

\bibitem{liu2023visual}
Liu, H., Li, C., Wu, Q., Lee, Y.J.: {Visual Instruction Tuning}. Advances in
  neural information processing systems  \textbf{36} (2024)

\bibitem{zhai2023sigmoid}
Zhai, X., Mustafa, B., Kolesnikov, A., Beyer, L.: {Sigmoid Loss for Language
  Image Pre-Training}. In: Proceedings of the IEEE/CVF International Conference
  on Computer Vision. pp. 11975--11986 (2023)

\bibitem{li2023blip2}
Li, J., Li, D., Savarese, S., Hoi, S.: {BLIP-2: Bootstrapping Language-Image
  Pre-training with Frozen Image Encoders and Large Language Models}. In:
  International conference on machine learning. pp. 19730--19742. PMLR (2023)

\bibitem{zhang2023faster}
Zhang, C., Han, D., Qiao, Y., et~al.: {Faster segment anything: Towards
  lightweight sam for mobile applications}. arXiv preprint arXiv:2306.14289
  (2023)

\bibitem{ilharco_gabriel_2021_5143773}
Ilharco, G., Wortsman, M., Wightman, R., et~al.: {OpenCLIP} (2021).
  \doi{10.5281/zenodo.5143773}

\bibitem{dosovitskiy2020image}
Dosovitskiy, A., Beyer, L., Kolesnikov, A., et~al.: {An image is worth 16x16
  words: Transformers for image recognition at scale}. arXiv preprint
  arXiv:2010.11929  (2020)

\bibitem{liu2022convnet}
Liu, Z., Mao, H., Wu, C.Y., et~al.: {A ConvNet for the 2020s}. In: Proceedings
  of the IEEE/CVF conference on computer vision and pattern recognition. pp.
  11976--11986 (2022)

\bibitem{chen2022pali}
Chen, X., Wang, X., Changpinyo, et~al.: Pali: A jointly-scaled multilingual
  language-image model. arXiv preprint arXiv:2209.06794  (2022)

\bibitem{wang2022image}
Wang, W., Bao, H., Dong, L., Bjorck, et~al.: {Image as a Foreign Language: BEiT
  Pretraining for All Vision and Vision-Language Tasks}. arXiv preprint
  arXiv:2208.10442  (2022)

\bibitem{he2024malmm}
He, B., Li, H., Jang, Y.K., et~al.: {MA-LMM: Memory-Augmented Large Multimodal
  Model for Long-Term Video Understanding}. arXiv preprint arXiv:2404.05726
  (2024)

\bibitem{schuhmann2022laion5b}
Schuhmann, C., Beaumont, R., Vencu, R., Gordon, C., Wightman, R., Cherti, M.,
  Coombes, T., Katta, A., Mullis, C., Wortsman, M., et~al.: {LAION-5B: An open
  large-scale dataset for training next generation image-text models}. Advances
  in Neural Information Processing Systems  \textbf{35},  25278--25294 (2022)

\bibitem{sanh2020distilbert}
Sanh, V., Debut, L., Chaumond, J., Wolf, T.: {DistilBERT, a distilled version
  of BERT: smaller, faster, cheaper and lighter}. arXiv preprint
  arXiv:1910.01108  (2019)

\bibitem{liu2019roberta}
Liu, Y., Ott, M., Goyal, et~al.: {RoBERTa: A Robustly Optimized BERT
  Pretraining Approach}. arXiv preprint arXiv:1907.11692  (2019)

\bibitem{8374622}
Chang, A., Dai, A., Funkhouser, T., et~al.: {Matterport3D: Learning from RGB-D
  Data in Indoor Environments}. In: 2017 International Conference on 3D Vision
  (3DV). pp. 667--676 (2017). \doi{10.1109/3DV.2017.00081}

\bibitem{xia2018gibson}
Xia, F., Zamir, A.R., He, Z., et~al.: {Gibson env: Real-world perception for
  embodied agents}. In: Proceedings of the IEEE conference on computer vision
  and pattern recognition. pp. 9068--9079 (2018)

\end{thebibliography}

\end{document}